\documentclass[conference,compsoc]{IEEEtran}
\usepackage{url,hyperref,microtype}
\usepackage{color}
\usepackage{float}
\usepackage{graphicx}
\usepackage[T1]{fontenc}
\usepackage{units}
\usepackage{verbatim}
\usepackage{mathtools}
\graphicspath{{img/}}

\ifCLASSOPTIONcompsoc
  \usepackage[nocompress]{cite}
\else
  \usepackage{cite}
\fi
\ifCLASSINFOpdf
\else
\fi
\begin{document}
%
\title{Conversion of Artificial Recurrent Neural Networks to Spiking Neural Networks for Low-power Neuromorphic Hardware}



%
\author{\IEEEauthorblockN{
Peter U. Diehl\IEEEauthorrefmark{1}$^{1}$,
Guido Zarrella\IEEEauthorrefmark{2}$^{1}$,
Andrew Cassidy\IEEEauthorrefmark{3}, 
Bruno U. Pedroni\IEEEauthorrefmark{4} and
Emre Neftci\IEEEauthorrefmark{4}\IEEEauthorrefmark{5} 
}
\IEEEauthorblockA{\IEEEauthorrefmark{1}Institute of Neuroinformatics\\
ETH Zurich and University Zurich, Switzerland\\ 
Email: peter.u.diehl@gmail.com}
\IEEEauthorblockA{\IEEEauthorrefmark{2}The MITRE Corporation, Bedford, MA, USA\\}
\IEEEauthorblockA{\IEEEauthorrefmark{3}IBM Research Almaden, San Jose, CA, USA\\}
\IEEEauthorblockA{\IEEEauthorrefmark{4}Institute for Neural Computation, UC San Diego, La Jolla, USA\\}
\IEEEauthorblockA{\IEEEauthorrefmark{5}Department of Cognitive Sciences, UC Irvine, Irvine, USA\\}
$^1$ Peter U. Diehl and Guido Zarrella have contributed equally to this work}


\maketitle

\begin{abstract}
In recent years the field of neuromorphic low-power systems gained significant momentum, spurring brain-inspired hardware systems which operate on principles that are fundamentally different from standard digital computers and thereby consuming orders of magnitude less power.
However, their wider use is still hindered by the lack of algorithms that can harness the strengths of such architectures.
While neuromorphic adaptations of representation learning algorithms are now emerging, the efficient processing of temporal sequences or variable length-inputs remain difficult, partly due to challenges in representing and configuring the dynamics of spiking neural networks. 
Recurrent neural networks (RNN) are widely used in machine learning to solve a variety of sequence learning tasks.
In this work we present a train-and-constrain methodology that enables the mapping of machine learned (Elman) RNNs on a substrate of spiking neurons, while being compatible with the capabilities of current and near-future neuromorphic systems.
This "train-and-constrain" method consists of first training RNNs using backpropagation through time, then discretizing the weights and finally converting them to spiking RNNs by matching the responses of artificial neurons with those of the spiking neurons.
We demonstrate our approach by mapping a natural language processing task (question classification), where we demonstrate the entire mapping process of the recurrent layer of the network on IBM's Neurosynaptic System "TrueNorth", a spike-based digital neuromorphic hardware architecture. 
TrueNorth imposes specific constraints on connectivity, neural and synaptic parameters.
To satisfy these constraints, it was necessary to discretize the synaptic weights to 16 levels, discretize the neural activities to 16 levels, and to limit fan-in to 64 inputs. 
Surprisingly, we find that short synaptic delays are sufficient to implement the dynamical (temporal) aspect of the RNN in the question classification task. 
Furthermore we observed that the discretization of the neural activities is beneficial to our train-and-constrain approach.
The hardware-constrained model achieved 74\% accuracy in question classification while using less than 0.025\% of the cores on one TrueNorth chip, resulting in an estimated power consumption of $\approx 17 \mu W$. 
\end{abstract}


%
\IEEEpeerreviewmaketitle

\section{Introduction}
The ever growing availability of large-scale neuromorphic hardware systems \cite{Merolla_etal14, Indiveri_etal06, Khan_etal08, Benjamin_etal14} enables large-scale simulations of neural networks in real-time on an extremely low power budget.
One interesting application of such neuromorphic systems is to use them for pattern recognition tasks such as handwritten digit recognition or natural language processing, which on the long-term could prove useful for example for mobile devices or robotic systems where energy efficiency is very important.

Currently, the best recognition performance of spiking networks on the most widespread machine learning benchmark MNIST (handwritten digit recognition) is based on a machine learning technique called convolutional neural network \cite{LeCun_etal98}. 
Those networks are pre-trained on a conventional computer and then converted to spiking neural networks (SNN) \cite{Diehl_etal15}.
Convolutional neural networks work extraordinarily well for vision  \cite{ciresan2010, krizhevsky2012, Sermanet_etal_2013} and auditory tasks  \cite{deng2013}). 
While convolutional neural networks have proven successful for some challenges in natural language processing  \cite{Kalchbrenner_etal_2014,Kim14}, the sequential nature of language lends itself to solutions that explicitly model histories of arbitrary length with complex dependencies across time. 

During the recent renaissance in machine learning neural network (NN) research, machine learning recurrent neural networks (RNN) have proven to be an essential tool for learning to interpret and generate language. 
RNNs have been trained to state-of-the-art performance on many challenging NLP tasks including language translation \cite{Bahdanau2014}, image caption generation \cite{Vinyals2015}, estimation of semantic similarity \cite{zarrella2015}, and language modeling \cite{kim2015}.
Note that despite also being called "recurrent neural network" this type of RNN is very different from the type of RNN typically seen in computational neuroscience and neuromorphic engineering (see discussion section for more details on other types of RNNs).
Therefore we will say explicitly if we refer to machine learning RNNs.

In this article, we extend the application of machine learning RNNs to the neuromorphic domain by converting them to spiking RNNs.
Specifically, the goal is to show \textit{how} to convert machine learning RNNs to spiking ones, for the purpose of simulating them on power-efficient hardware while maintaining high classification performance. 
Equipped with this approach, future advances in the development of machine learning RNNs can lead to higher performing spiking RNNs.
Our workflow for solving this conversion task consists of first training Elman RNNs (a simple machine learning RNN) \cite{elman1990} on a conventional computer and then using the trained weights with the defined connectivity to create a spiking equivalent. 
This spike-based RNN is then implemented on IBM's Neurosynaptic System "TrueNorth".

Some of the issues that arise during the conversion of Elman recurrent networks to spiking neural networks have been addressed by the conversion of convolutional neural networks and fully-connected networks, e.g. the substitution of artificial neurons with spiking neurons by using rectified linear units (ReLU) and replacing them with integrate-and-fire neurons \cite{Diehl_etal15, Diehl_etal16a, Cao_etal14}.
However, RNNs  feed back the activity of the hidden-neurons to themselves. 
This is a distinguishing feature of RNNs that serves as a memory of previous inputs. 
In spiking RNNs, this feedback must be represented in a spiking fashion and presents one of the most challenging aspects of spike-based RNNs.
Surprisingly, we find that using synaptic delays lasting only 15 time steps, effectively corresponding to a 4 bit discretization of the hidden-state, does not impair the functionality of the RNN.

While it already represents a challenge to design high-performance spike-based recurrent networks, implementing those on neuromorphic hardware adds another layer of complexity.
Most neuromorphic systems pose several constraints on the types of networks that can be implemented like limited connectivity or limited resolution of the synaptic weights.
Therefore only a few recognition systems have been implemented on neuromorphic hardware \cite{Neil_Liu14, stromatias2015,Esser_etal2015}.

\section{Material \& Methods}
In this section we describe the chronological process of creating a spike-based RNN.
We start by explaining the exact task and the associated dataset, then describe the pre-processing of the data and after that the architecture of the machine learning RNN.
All parts described in these subsections (sections \ref{sec:task} up to \ref{sec:architecture}) are, with some modifications, based on existing work.
The following subsection \ref{sec:trueNorth} describes the main contribution of this work (besides the introduction to TrueNorth itself).
We explain how the input is converted from rates to spikes and how the spikes are converted back to rates to calculate the output.
After that we cover the discretization of the trained weights for use on TrueNorth.
The last (and most important) part of the TrueNorth subsection covers how the hidden state of the recurrent layer is represented on TrueNorth. 
Finally, the last subsection covers two different versions of the machine learning RNN that are used for comparison to our TrueNorth implementation.

\subsection{Question Classification Task}\label{sec:task}
Here we use the question classification data set presented in \cite{Li_Roth02}.
The goal of this task is to classify question sentences into one of six coarse categories and potentially into finer grained subcategories. 
For example, the question: "How much does the human adult female brain weigh ?" expects an answer of the type "Number". 
In the previous example, "Number" is the coarse category, while "fine" type is the "weight" category. 
In this project we used only the coarse categories which included: Abbreviation, Description, Number, Entity, Human, and Location. 
The training dataset consisted of 5000 labeled sentences, and a evaluation (test) set that is not used during training consisting of 500 labeled sentences.
For training and testing of the recurrent network we also added a special End Of Sentence (EOS) word (a vector of zeros) at the end of every input sentence. 

\subsection{Pre-Processing and Word Vectors}
We aim to train our system to generalize from a narrow training set to the entire universe of possible questions. 
It must be robust to many forms of natural linguistic variation that may occur. 
For example, if the question above were instead submitted as "What is the mass of an average man's brain ?" the system would benefit from understanding  that mass and weight have similar attributes. 
Therefore we equip our algorithm with a model of word semantics trained in advance using the word2vec \cite{mikolov2013} library. 
Word vectors such as these have been employed in many state-of-the-art language processing systems in both academia and industry \cite{levy2014, li2015}.

The goal of word2vec is to embed each word into a high-dimensional space such that words with latent semantic similarity ("good", "awesome") are near each other but distant from dissimilar words ("terrorist", "aardvark").
We used the word2vec skipgram variant with negative sampling, which effectively learns to model word meaning by predicting word coocurrences from a large text corpus.
Specifically, we trained 64-dimensional word vectors using 3.4 billion tokens from text of the English Wikipedia.
The preprocessing included removing punctuation, setting all letters to lower case, and substituting uncommon words with an 'unknown' token. 
Our training process resulted in vector representations for 324264 commonly occurring words.
Input questions were then transformed to the sequence of their word vectors, with one vector for each word in the input, and with unknown words mapped to the average word vector.

\begin{figure}
\begin{center}
\includegraphics[width=8.5cm]{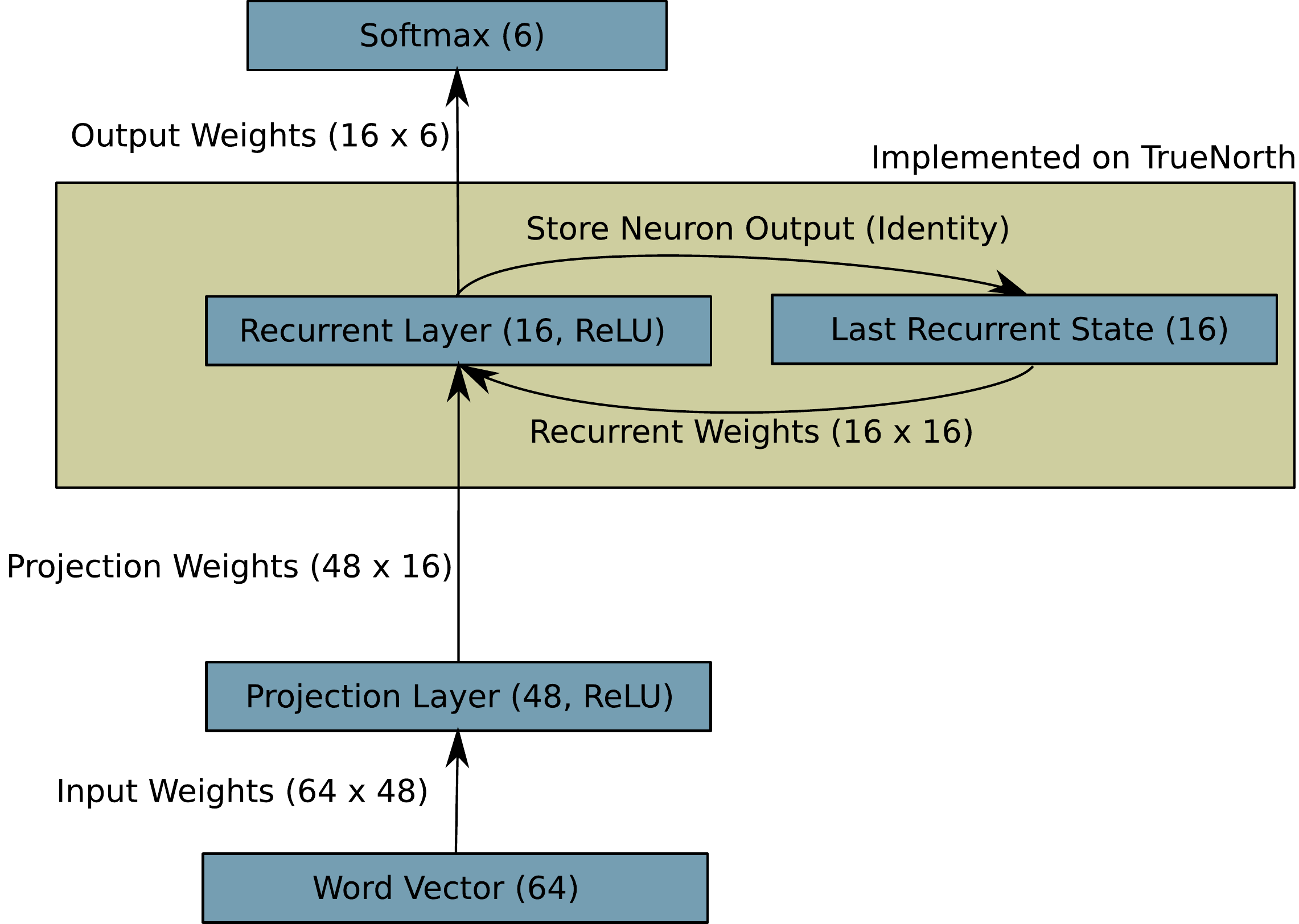}
\end{center}
 \caption{ \emph{Recurrent Neural Network model for solving the question classification task.} The network consisted of a projection layer (48 units), a recurrent layer (16 units) and a softmax layer for classification (6 units). At each time step, the recurrent layer takes an input from the projection layer and the previous step of the recurrent layer. The output the recurrent state is used for classification. The scope of our implementation is indicated by the shaded box. }\label{fig:01}
\end{figure}

\subsection{Neural Network Architecture and Training}\label{sec:architecture}
The machine learning RNN we train as a basis for the spiking RNN exclusively uses standard techniques from machine learning.
Besides using ReLUs there are no special requirements for the architecture itself and can therefore be modified easily, for example to use convolutional features or a deeper or bigger network to achieve better performance.

The network consists of a projection layer (48 units), a recurrent layer (16 units) and a softmax layer for classification (6 units), see figure \ref{fig:01}. 
This combination of different types of layers, here a so called fully-connected (or projection) layer, an Elman or simple RNN and a softmax layer is common in machine learning NNs \cite{schmidhuber2015}.
The dimension of the projection layer and the recurrent layer were constrained to fit on one core of a TrueNorth chip (see below). 
Furthermore, due to good performance and to ease the mapping to TrueNorth spiking neurons, the network utilized rectified linear units (ReLU) without biases \cite{zeiler2013} throughout.
The neural network was trained using "backpropagation through time" with stochastic gradient descent \cite{werbos1990}.
This was trained using the Pylearn2 and Theano packages \cite{Bergstra_etal10}.

All of the 48 neurons in the projection layer receive the 64-dimensional word vector as inputs.
The output of those 48 neurons are then used as 48-dimensional input for each one of 16 neurons in the recurrent layer.
Additionally, each neuron in the recurrent layer receives input from all 16 neurons in the recurrent layer, hence the name recurrent.
The output of the recurrent layer neurons are then fed as input to a softmax layer that computes the final classification.

\subsection{TrueNorth Implementation}\label{sec:trueNorth}
\subsubsection{TrueNorth and Neuron Conversion}
The IBM Neurosynaptic System "TrueNorth" is a non-von Neumann architecture that integrates 1 million programmable spiking neurons \cite{Merolla_etal14}. 
The system consists of 4096 cores with 256 neurons per core, and each core can accommodate up to 65536 synapses in a crossbar fashion. 
Each neuron's equations and synaptic states are updated every millisecond, which we will call 1 tick.
Note that although a continuous time neural network is being simulated, those discrete updates are a common way to numerically approximate the change of a continuous system.

\begin{figure}
\begin{center}
\includegraphics[width=8.5cm]{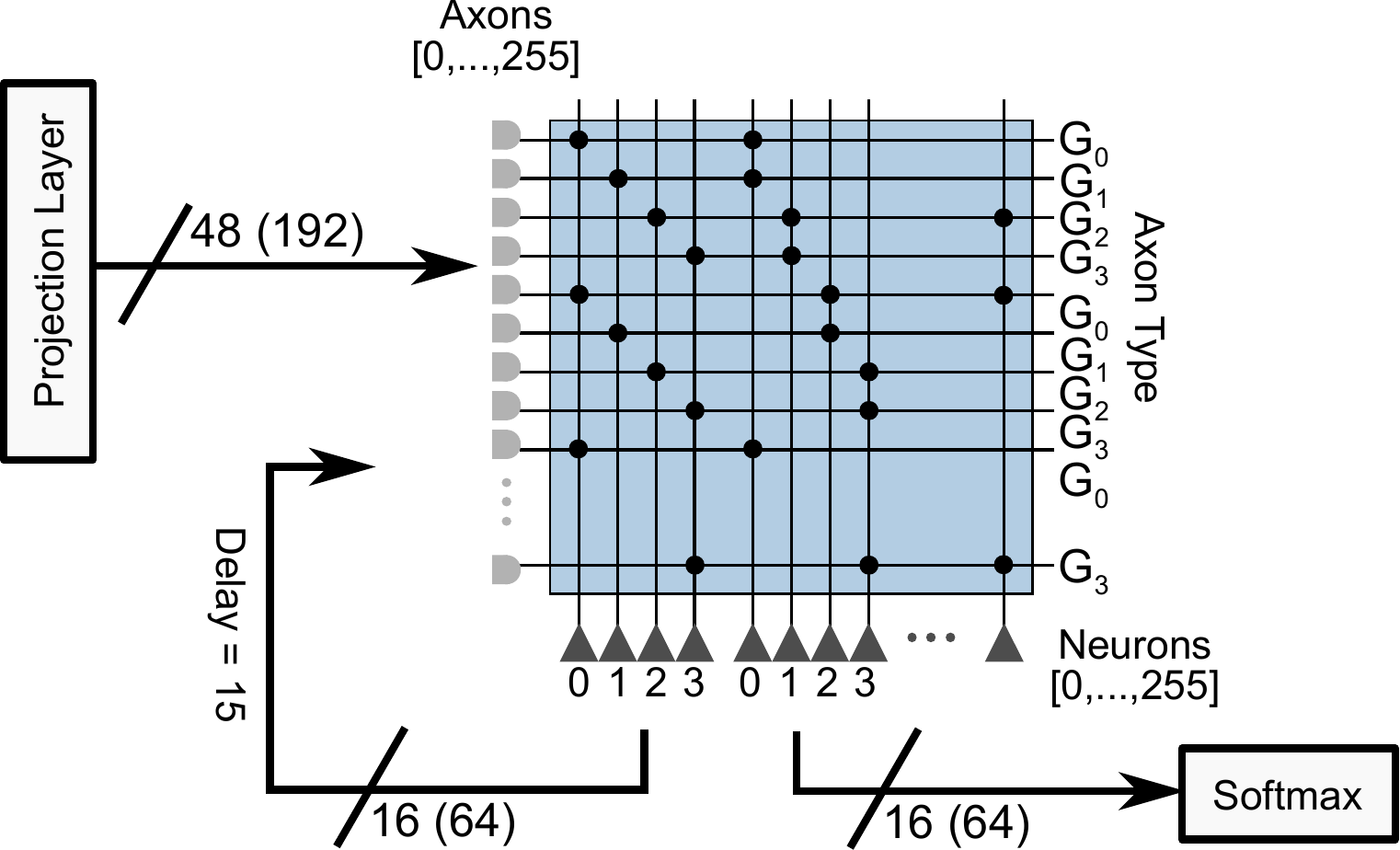}
\end{center}
\caption{  \emph{TrueNorth implementation of a recurrent neural network.} The feedback in the recurrent network is implemented using the synaptic delays (delay of 15 ticks, the maximum supported on TrueNorth). The effect of the maximum delay is to limit to 16 ticks the time window during which the spikes are counted (15 ticks from the delay plus 1 tick for transmission). For every connection, we used 4 axons per dimension to implement 4-bit weights. The input to the recurrent neural network and the output to the softmax classifier is computed offline on the computer. This architecture can support recurrent neural networks that verify the condition $N_{in} + N_{hid} <= 256/N_s$. }\label{fig:TNcore}
\end{figure}

By using four synapses per actual input, 4-bit precision synapses can be implemented, which leads to a fan-in of 64 per neuron. 
As a proof of concept, we focused on implementing the recurrent layer on TrueNorth, see figure \ref{fig:TNcore}.
We only used a single core since it is sufficient to understand the conversion method and the main challenge when converting RNNs compared to other networks like convolutional neural networks.
We chose to only convert the recurrent layer (i.e. the part with solid blue background in figure \ref{fig:01} or its TrueNorth implementation in figure \ref{fig:TNcore}) and not convert the other layers since those have been successfully implemented on TrueNorth in the past \cite{Esser_etal2015,Diehl_etal15, Diehl_etal16a}.

Using $N_s$-bit weights, the dimension of the recurrent layer is limited to $ N_{in}+N_{hid} \le 256/N_s$, where $N_{in}$ is the number of inputs, $N_{hid}$ is the number of (hidden) units in the recurrent layer and $N_s$ is the resolution of the connection weights in bits.

To map the ReLUs onto TrueNorth, we used linear neurons, whose membrane state follows the following dynamics:
\begin{equation}
\begin{split}
    V(t+1) &= V(t) + \sum_{i=0}^{255} A_i w_{ij} s_j^{G_i} \\
\text{if }&V(t)<0: V(t) \leftarrow 0 \\
\text{if }&V(t)>T: V(t) \leftarrow V(t) - T
\end{split}
\end{equation}

where $A_i(t)$ is the input spike on axon i, $s_j^{G_i}$ is the synaptic weight, $G_i \in \{0, 1, 2, 4\}$ is the axon type and $w_{ij} \in {0,1}$ is the synaptic connectivity between axon $i$ and neuron $j$. 
For the full neuron equation see \cite{cassidy2013}.
The above equations imply that, after a spike is elicited, the membrane state is subtracted an amount corresponding to the firing threshold. 
This enables the neuron to fire a number of times that is proportional to the synaptic input.

The model was written in MATLAB, using the integrated programming environment for IBM's Neurosynaptic System. One parameter of the model is the simulation platform. By switching this parameter between "TN" and "NSCS" the exact same program can be run on a connected TrueNorth chip or it can be run using the NSCS simulation environment. Note that there is an exact one to one correspondence between the results of the simulation environment and the TrueNorth chip, which means the code can be run using either system (given that the user has a TrueNorth chip available).

\subsubsection{Input Encoding and Output Decoding}
During the training of the Elman RNN, the input to the recurrent layer is encoded by the output of the projection layer.
However, after conversion to a spiking network, the input needs to be provided in the form of spikes.
Here we use a simple rate code of the output, i.e. the higher the input of the recurrent layer the higher the number of input spikes.
More specifically, we use Poisson spike trains with firing rates corresponding to the rate of the represented input dimension of the projection layer. 
Each word in an example sentence is presented for 16 ticks.
This means there can be up to 16 input spikes for each of the 48 inputs and for each input word.
Therefore the input resolution is discretized from 32 bit to 4 bit.
Between different sentences we reset the neuron and synapse state.

Similarly, we use a rate code for the output of the recurrent layer, i.e. each of the 16 neurons can fire up to 16 spikes which again represents a discretization of the 32 bit precision used for training to 4 bit.

\subsubsection{Weight Discretization}
The weight discretization method described here is the same as described in \cite{Diehl_etal16a}.
Since each synapse on TrueNorth is either present or not but its weight can only be one of the 4 chosen types, the weight resolution could be interpreted as being single bit.
However, by using 4 axons for each actual input, it is possible to achieve a 4-bit accuracy for each actual input.
For example, by choosing the 4 axon types to be \{1,2,4,-8\} and then combining them appropriately any number between -8 and 7 can be represented.

To accommodate the network parameters on TrueNorth, the weights of the machine learning RNN were bounded to (-1, 1), scaled and discretized to 4 bit. 
To cancel the effect of the scaling, the input to the recurrent layer was scaled by 1/16.

\begin{figure}
\begin{center}
\includegraphics[width=17cm]{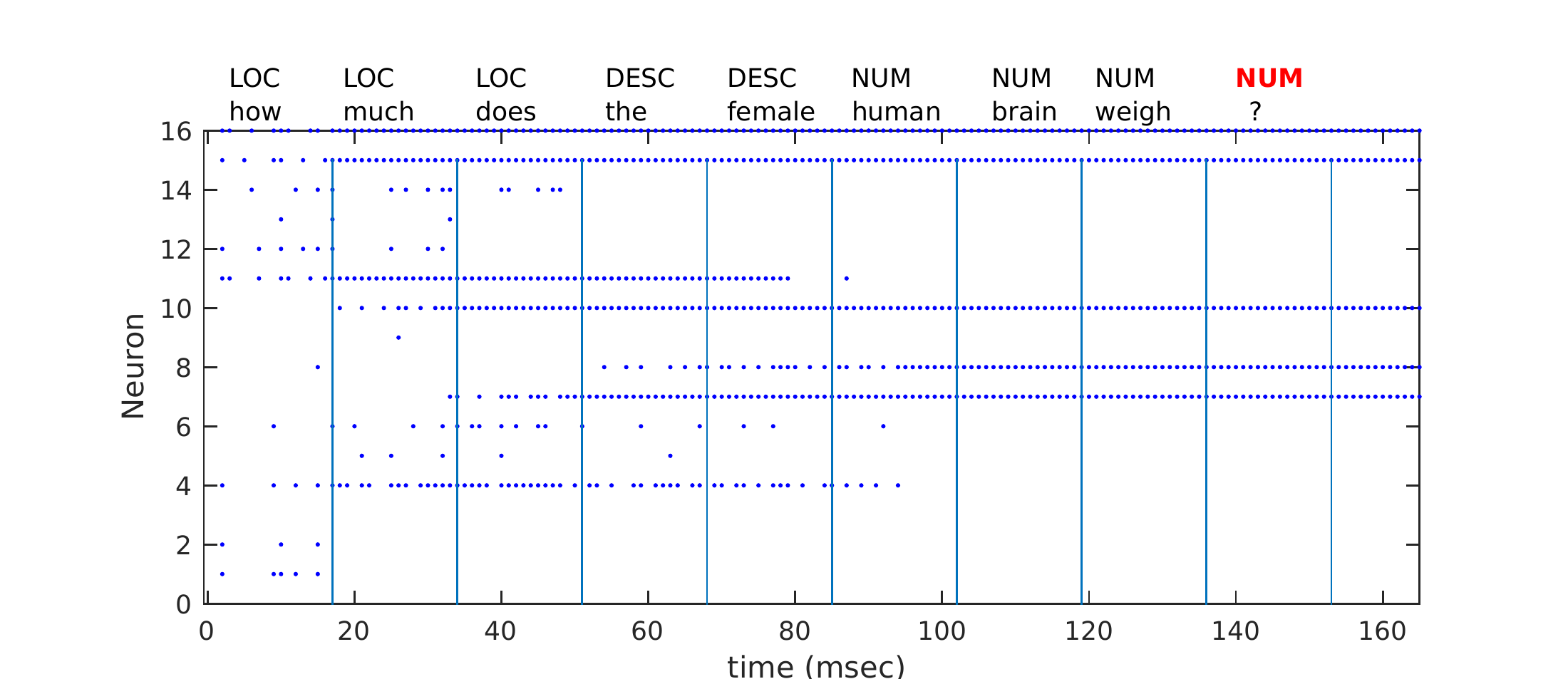}
\end{center}
\caption{ \emph{Sample Raster Plot of the TrueNorth for the recurrent units.} The question class for the question "How much does the female brain weigh ? " is correctly identified as "Number" (NUM). Note that the shown neurons are located in the recurrent layer and do not represent the output neurons. The shown 16 neurons are combine in the output layer to determine the class with the highest response.}\label{fig:sample_run}
\end{figure}

\subsubsection{Hidden State Encoding}
The encoding of the hidden state represents a crucial part of this work since this is (besides the different training for the Elman RNN) the main difference to fully-connected networks.
The challenge is that the hidden state in the Elman RNN is the output of the recurrent layer for the \textit{last} input word.
While this is easy to implement on traditional hardware, it is not immediately clear how to "store" this for the next input.

We decided to represent the hidden state of the spiking RNN using synaptic delays.
Using a single connection, True North has the ability to implement delays up to 15 ticks, i.e. the spike will arrive 16 ticks (1 tick from transmission and 15 ticks from delay) after the source neuron fired at the destination neuron.
Therefore, 16 ticks of the spiking RNN correspond to one time step in the machine learning RNN.
Note that this is the reason we chose a duration of 16 ticks for each input word.

An example time course is depicted in figure \ref{fig:sample_run}.
At ticks 1 to 16, the input spikes from the projection layer to the recurrent layer represent the word "how".
There are no other spikes arriving during those first 16 ticks.
However, as soon as one of the neurons in the recurrent layer spikes, all recurrent neurons which are connected to this neuron will receive this spike exactly 16 ticks later, e.g. the spike fired by neuron 13 at tick 10 which means that all neurons connected to neuron 13 will receive a spike at tick 26 (in addition to other possible spikes from recurrent neurons and in addition to the input spikes from the projection layer).
Note that the spikes are not being aggregated but instead arrive always with the exact time difference they were fired in.
This is an important difference to the machine learning RNN where all information is aggregated before a result is calculated and which potentially leads to decreased perfomance of the spiking RNN compared to the machine learning RNN.

After the first word was presented, the spikes corresponding to the next word "much" are used as input from the projection layer during the ticks 17 to 32.
Simultaneously, the spikes that were fired by the neurons from the recurrent layer during the first word "how" are arriving.
Therefore the delay lines are essentially storing the spikes fired during the last input word.

The integration time effectively determines the discretization of the activation function or how many spikes can be fired for each example.
Since the duration of the delay has to be equal to the time intervals between new inputs, the maximum delay limits the possible integration time.
Note that it is possible to increase this delay by chaining delays of additional axons (16 more ticks per axon).

\subsection{Setups for Comparison}
In order to be able to better compare the performance of the machine learning RNN with its TrueNorth counterpart, we used two intermediate setups.
The first one is equivalent to the original machine learning RNN but the weights are scaled and discretized to 4 bit (since the ReLUs have no bias, the scaling actually has no influence on the performance).
The second setup uses, in addition to the weight discretization to 4 bit, a discretization of the hidden state to 4 bit.
This is achieved by discretizing the ReLU function such that the activation is one of 16 different values to mimic the TrueNorth neuron.

\section{Results}
For all four setups we used the question classification test set introduced in \cite{Li_Roth02}.
The respective accuracy of all four setups is shown in table \ref{tab:results}.
Training of the original ReLU NN with floating point weights yields a classification accuracy of 85\%.
The variance of the results was obtained by using different initializations of the parameters of the original network.
When reducing the precision of the weights to 4 bit, the accuracy dropped to 72.2\%. 
In the next step we discretized the hidden state to 4 bit.
To our surprise, this modification increased the accuracy of the network to 78.4\%.

Lastly, the network was implemented as TrueNorth network by substituting the ReLUs by TrueNorth linear neurons and converting the 48-dimensional real-valued inputs to 48 Poisson spike-trains, each with a firing-rate proportional to the values of the corresponding input dimension. 
The resulting TrueNorth network shows an accuracy of 74\% on the question classification test set.

\begin{table*}\label{tab:results}
\begin{center}
\begin{tabular}{ll}\hline
\textbf{Configuration} &	\textbf{Accuracy} \\
ReLUs, 32bit weights, 32bit hidden state (PC) & 85\%\\
ReLUs, scaled 4bit weights 32bit hidden state (PC) &	72.2\% \\
ReLUs, scaled 4bit weights, 4bit hidden state (PC) &	78.4\% \\
TrueNorth neurons, scaled 4bit weights, 4bit spiking hidden state (TrueNorth) &	74\% \\ \vspace{9pt}
\end{tabular}\caption{Accuracy of the different networks}
\end{center}
\end{table*}

We also embedded this TrueNorth network in an interactive question classification system, where a user can type in a question and the system outputs the question type as well as a plot of the spike-responses of the hidden neurons.
Besides being able to classify more obvious questions like "where was peter born" as "location", it can also deal with more ambiguous cases where the question word does not determine the question type.
For example it classifies "what city was peter born in" also as a "location", "what is the meaning of life" as  "description", and "what is the company that created truenorth" as "entity".
Since 'peter' is a word that often occurs in Wikipedia, it is represented in the word space and can therefore be part of a query.
On the other hand, the word 'truenorth' is not often mentioned in the Wikipedia corpus and is therefore not represented in the word space.
In our interactive system we substitute such words with the average over all word vectors to minimize distortion.

\section{Discussion}
\subsection{Result Interpretation}
Our results demonstrate a proof-of-concept recurrent neural network that can be trained offline and afterwards mapped onto the highly power-efficient TrueNorth chip. 
Furthermore, we show that synaptic delays are sufficient for supporting the temporal dynamics of simple recurrent neural networks.
Using a 15 tick delay for "storing" the state of the neurons corresponds to discretizing the state to 4 bit.
However, while the accuracy of the machine learning RNN is comparable to reported results in the original and following studies (84\% - \cite{Li_Roth02}, 86.2\% - \cite{krishnan2005} and 85.6\% - \cite{huang2008}), there is a performance gap between the machine learning RNN and the TrueNorth network.
By having a closer look at the four different models we can understand why this gap exists.

The biggest drop in performance is due to the discretization of the synaptic weights, as can be seen by comparing the first and the second model in table \ref{tab:results}.
However, this performance decrease due to discretization is expected and has been the topic of other studies.
In order to reduce performance losses due to weight discretization, it is possible to choose better discretization methods than simple rounding.
For example, it is possible to include discretization in the training of the network \cite{stromatias2015} (by using rounded weights during the forward pass of backpropagation) or by rounding probabilistically after training \cite{Muller_Indiveri15}. 

The second step was to discretize the hidden state to 4 bit.
Interestingly, this discretization does not decrease the classification performance of the network but rather increases it.
While we do not assume that this holds true in general for RNNs, the discretization might help in cases where there is only a limited number of target labels (in the presented example it was six).
A possible explanation is that the discretization prevents the drift between states if not much relevant information is contained in the new input.
As an example, if a questions starts with "where", it is highly likely that the answer is a location.
This "location" state of the hidden neurons then needs to be maintained in the face of irrelevant information like "is" or "the" which is especially challenging for recurrent networks for longer input sequences. 

Lastly, the mapping and the corresponding conversion to a spiking network caused a drop in performance of 4.4\%.
While this is much less than the drop caused by the weight discretization, it is still significant.
However, other studies that investigated the performance loss due to conversion from ReLUs to spiking units shows that the goes to zero as integration times increase, see \cite{Diehl_etal15, Diehl_etal16a}.
This increase in integration time is very easy to achieve in systems that do not rely on feedback from a recurrent layer by simply presenting each example for a longer period.
When using the delay mechanism presented here it is also necessary to be able to use delays which are as long as the desired integration time. 
One possibility would be to use multiple axon in a chain (of course this would come at the expense of available axons and it might be more useful to instead increase the size of the network).

\subsection{Comparison to Recurrent Neural Networks}
In this work we used established machine-learning techniques, specifically Elman RNNs \cite{elman1990}, and converted them to spiking neural networks.
The reason for using this approach is that our goal is to achieve competitive performance on practically relevant tasks and similar conversion methods already have been shown to be very effective for vision tasks \cite{Diehl_etal15, Cao_etal14}. 
However, more biologically plausible recurrent neural networks have been used for decades \cite{Wilson_Cowan72}.
In contrast to machine-learning neural networks, learning more biological neural networks for real-world pattern recognition tasks only started to gain momentum recently \cite{Habenschuss_etal13, Kappel_etal14, Neftci_etal13a, Neftci_etal14, Diehl_Cook15}.
Such networks are usually trained using spike-timing-dependent plasticity (STDP) \cite{Abbott_Song99} but there are also studies suggesting the usage of morphological learning \cite{Hussain_etal14}.
At this point however, more biologically realistic approaches are not (yet) competitive with deep neural network to SNN conversion methods (in terms of recognition performance).
Specifically, MNIST which is the most common benchmark in machine learning, the best reported performance using conversion methods is above 99\% \cite{Diehl_etal15} and the best performance using more biologically plausible networks is 95\% \cite{Diehl_Cook15}.

The decision to use Elman networks instead of more sophisticated machine learning RNNs like long short term memory networks (LSTM) \cite{Hochreiter_Schmidhuber97} or gated recurrent units (GRU) \cite{cho2014} was based on the short length of the input sequences, which allowed us to train without a significant risk of vanishing gradients. 
Additionally, recently Elman networks have been shown to match the performance of state-of-the-art LSTMs and GRUs when using some improvements on the basic structure such as restricting the recurrent connectivity matrix \cite{mikolov2014} and correctly initializing the recurrent connectivity matrix \cite{le2015}.
Since we are focusing on the conversion and not optimizing the performance of the machine learning RNN, we did not use those methods here.
Nor did we spent significant time optimizing our word vectors to maximize performance on this particular task.
However, the advantage of the conversion method is that such modifications can be applied to the underlying machine learning RNN without needing to change any parts the conversion method.
Moreover, other improvements (possibly found in the years to come) in training procedures, weight initialization or better loss functions are equally easy to use in conjunction with the presented framework since also all of them require no changes in the presented conversion method.

\subsection{Implications}
We showed how to implement high-performance recurrent neural networks on neuromorphic hardware. 
This offers great potential for a range of extremely low-power natural language understanding applications.
While the presented conversion method was only tested for a question classification task, the underlying machine learning RNN has been successfully used in a wide variety of setups including vision, audio and natural language processing tasks \cite{mnih2014, graves2013, boulanger2013, mikolov2012}.
Since the presented conversion method is oblivious to which task the machine learning RNN was trained on, it can be easily applied to other domains and tasks.
While traditional hardware on computers such as CPU's or GPU's easily consume 100 $W$ or more, an entire TrueNorth chip consumes up to 70 $mW$ and contains 4096 cores \cite{Merolla_etal14}.
Here, we only used one of those cores, which means that our system has an estimated power budget of only about 17 $\mu W$. 
Even assuming that there is overhead in using only one core, there are orders of magnitudes difference in power consumption between the used neuromorphic system (as there would be with almost all other neuromorphic systems) and traditional hardware.

Considering this extremely low power-budget, such pattern recognition systems can prove useful for low-power applications like mobile devices or robotics but also for server farms where power consumption presents a major cost factor.
In future work it would be interesting to see how the discretization of the hidden state influences the performance for more complex tasks, e.g. when predicting more than one of six classes.
Another point to be addressed is to better understand the influence the discretization of the hidden state has on the hidden state trajectory over time, i.e. how it perturbs the occurring changes.

\section*{Acknowledgments}
We thank the organizers and the participants of the Telluride Neuromorphic Cognition Engineering Workshop 2015, and especially the natural language processing group and Rodrigo Alvarez, John Arthur, Paul Merolla for fruitful discussions and the stimulating working environment.
We also thank the reviewers for their very helpful comments.

\textit{Funding:} 
PUD: SNF Grant 200021-143337 "Adaptive Relational Networks."
BUP: The Office of Naval Research (ONR MURI 14-13-1-0205) and CNPQ Brazil (CsF 201174/2012-0)
EN: the Office of Naval Research (ONR MURI 14-13-1-0205)




%

\bibliographystyle{IEEEtran}
\bibliography{biblio}

\begin{thebibliography}{10}
\providecommand{\url}[1]{#1}
\csname url@samestyle\endcsname
\providecommand{\newblock}{\relax}
\providecommand{\bibinfo}[2]{#2}
\providecommand{\BIBentrySTDinterwordspacing}{\spaceskip=0pt\relax}
\providecommand{\BIBentryALTinterwordstretchfactor}{4}
\providecommand{\BIBentryALTinterwordspacing}{\spaceskip=\fontdimen2\font plus
\BIBentryALTinterwordstretchfactor\fontdimen3\font minus
  \fontdimen4\font\relax}
\providecommand{\BIBforeignlanguage}[2]{{%
\expandafter\ifx\csname l@#1\endcsname\relax
\typeout{** WARNING: IEEEtran.bst: No hyphenation pattern has been}%
\typeout{** loaded for the language `#1'. Using the pattern for}%
\typeout{** the default language instead.}%
\else
\language=\csname l@#1\endcsname
\fi
#2}}
\providecommand{\BIBdecl}{\relax}
\BIBdecl

\bibitem{Merolla_etal14}
P.~A. Merolla, J.~V. Arthur, R.~Alvarez-Icaza, A.~S. Cassidy, J.~Sawada,
  F.~Akopyan, B.~L. Jackson, N.~Imam, C.~Guo, Y.~Nakamura \emph{et~al.}, ``A
  million spiking-neuron integrated circuit with a scalable communication
  network and interface,'' \emph{Science}, vol. 345, no. 6197, pp. 668--673,
  2014.

\bibitem{Indiveri_etal06}
\BIBentryALTinterwordspacing
G.~Indiveri, E.~Chicca, and R.~Douglas, ``A {VLSI} array of low-power spiking
  neurons and bistable synapses with spike--timing dependent plasticity,''
  \emph{{IEEE} Transactions on Neural Networks}, vol.~17, no.~1, pp. 211--221,
  Jan 2006. [Online]. Available:
  \url{http://ncs.ethz.ch/pubs/pdf/Indiveri_etal06.pdf}
\BIBentrySTDinterwordspacing

\bibitem{Khan_etal08}
M.~Khan, D.~Lester, L.~Plana, A.~Rast, X.~Jin, E.~Painkras, and S.~Furber,
  ``Spinnaker: mapping neural networks onto a massively-parallel chip
  multiprocessor,'' in \emph{Neural Networks, 2008. IJCNN 2008.(IEEE World
  Congress on Computational Intelligence). {IEEE} International Joint
  Conference on}.\hskip 1em plus 0.5em minus 0.4em\relax IEEE, 2008, pp.
  2849--2856.

\bibitem{Benjamin_etal14}
B.~V. Benjamin, P.~Gao, E.~McQuinn, S.~Choudhary, A.~R. Chandrasekaran,
  J.~Bussat, R.~Alvarez-Icaza, J.~V. Arthur, P.~Merolla, and K.~Boahen,
  ``Neurogrid: A mixed-analog-digital multichip system for large-scale neural
  simulations,'' \emph{Proceedings of the IEEE}, vol. 102, no.~5, pp. 699--716,
  2014.

\bibitem{LeCun_etal98}
Y.~LeCun, L.~Bottou, Y.~Bengio, and P.~Haffner, ``Gradient-based learning
  applied to document recognition,'' \emph{Proceedings of the IEEE}, vol.~86,
  no.~11, pp. 2278--2324, 1998.

\bibitem{Diehl_etal15}
P.~U. Diehl, D.~Neil, J.~Binas, M.~Cook, S.-C. Liu, and M.~Pfeiffer,
  ``Fast-classifying, high-accuracy spiking deep networks through weight and
  threshold balancing,'' in \emph{International Joint Conference on Neural
  Networks (IJCNN),}.\hskip 1em plus 0.5em minus 0.4em\relax IEEE, 2015, pp.
  1--8.

\bibitem{ciresan2010}
D.~C. Ciresan, U.~Meier, L.~M. Gambardella, and J.~Schmidhuber, ``Deep, big,
  simple neural nets for handwritten digit recognition,'' \emph{Neural
  Computation}, vol.~22, no.~12, pp. 3207--3220, 2010.

\bibitem{krizhevsky2012}
A.~Krizhevsky, I.~Sutskever, and G.~E. Hinton, ``Imagenet classification with
  deep convolutional neural networks,'' in \emph{Proc. of NIPS}, 2012, pp.
  1097--1105.

\bibitem{Sermanet_etal_2013}
P.~Sermanet, D.~Eigen, X.~Zhang, M.~Mathieu, R.~Fergus, and Y.~LeCun,
  ``{OverFeat}: Integrated recognition, localization and detection using
  convolutional networks,'' \emph{arXiv preprint}, vol. 312.6229, 2013.

\bibitem{deng2013}
L.~Deng, G.~Hinton, and B.~Kingsbury, ``New types of deep neural network
  learning for speech recognition and related applications: An overview,'' in
  \emph{Acoustics, Speech and Signal Processing (ICASSP), 2013 IEEE
  International Conference on}.\hskip 1em plus 0.5em minus 0.4em\relax IEEE,
  2013, pp. 8599--8603.

\bibitem{Kalchbrenner_etal_2014}
\BIBentryALTinterwordspacing
N.~Kalchbrenner, E.~Grefenstette, and P.~Blunsom, ``A convolutional neural
  network for modelling sentences,'' \emph{Proceedings of the 52nd Annual
  Meeting of the Association for Computational Linguistics}, June 2014.
  [Online]. Available: \url{http://goo.gl/EsQCuC}
\BIBentrySTDinterwordspacing

\bibitem{Kim14}
\BIBentryALTinterwordspacing
Y.~Kim, ``Convolutional neural networks for sentence classification,''
  \emph{CoRR}, vol. abs/1408.5882, 2014. [Online]. Available:
  \url{http://arxiv.org/abs/1408.5882}
\BIBentrySTDinterwordspacing

\bibitem{Bahdanau2014}
\BIBentryALTinterwordspacing
D.~Bahdanau, K.~Cho, and Y.~Bengio, ``Neural machine translation by jointly
  learning to align and translate,'' \emph{CoRR}, vol. abs/1409.0473, 2014.
  [Online]. Available: \url{http://arxiv.org/abs/1409.0473}
\BIBentrySTDinterwordspacing

\bibitem{Vinyals2015}
O.~Vinyals, A.~Toshev, S.~Bengio, and D.~Erhan, ``Show and tell: A neural image
  caption generator,'' in \emph{The IEEE Conference on Computer Vision and
  Pattern Recognition (CVPR)}, June 2015.

\bibitem{zarrella2015}
G.~Zarrella, J.~Henderson, E.~M. Merkhofer, and L.~Strickhart, ``Mitre: Seven
  systems for semantic similarity in tweets,'' \emph{Proceedings of SemEval},
  2015.

\bibitem{kim2015}
\BIBentryALTinterwordspacing
Y.~Kim, Y.~Jernite, D.~Sontag, and A.~M. Rush, ``Character-aware neural
  language models,'' \emph{CoRR}, vol. abs/1508.06615, 2015. [Online].
  Available: \url{http://arxiv.org/abs/1508.06615}
\BIBentrySTDinterwordspacing

\bibitem{elman1990}
J.~L. Elman, ``Finding structure in time,'' \emph{Cognitive science}, vol.~14,
  no.~2, pp. 179--211, 1990.

\bibitem{Diehl_etal16a}
P.~U. Diehl, B.~Pedroni, A.~Cassidy, P.~Merolla, E.~Neftci, and G.~Zarrella,
  ``Truehappiness: Sentiment analysis on truenorth,'' \emph{arXiv}, 2016.

\bibitem{Cao_etal14}
Y.~Cao, Y.~Chen, and D.~Khosla, ``Spiking deep convolutional neural networks
  for energy-efficient object recognition,'' \emph{International Journal of
  Computer Vision}, pp. 1--13, 2014.

\bibitem{Neil_Liu14}
D.~Neil and S.-C. Liu, ``Minitaur, an event-driven fpga-based spiking network
  accelerator,'' \emph{Very Large Scale Integration (VLSI) Systems, IEEE
  Transactions on}, vol.~22, no.~12, pp. 2621--2628, 2014.

\bibitem{stromatias2015}
E.~Stromatias, D.~Neil, M.~Pfeiffer, F.~Galluppi, S.~B. Furber, and S.-C. Liu,
  ``Robustness of spiking deep belief networks to noise and reduced bit
  precision of neuro-inspired hardware platforms,'' \emph{Frontiers in
  neuroscience}, vol.~9, 2015.

\bibitem{Esser_etal2015}
S.~K. Esser, R.~Appuswamy, P.~Merolla, J.~V. Arthur, and D.~S. Modha,
  ``Backpropagation for energy-efficient neuromorphic computing,'' in
  \emph{Advances in Neural Information Processing Systems}, 2015, pp.
  1117--1125.

\bibitem{Li_Roth02}
X.~Li and D.~Roth, ``Learning question classifiers,'' in \emph{Proceedings of
  the 19th international conference on Computational linguistics-Volume
  1}.\hskip 1em plus 0.5em minus 0.4em\relax Association for Computational
  Linguistics, 2002, pp. 1--7.

\bibitem{mikolov2013}
T.~Mikolov, K.~Chen, G.~Corrado, and J.~Dean, ``Efficient estimation of word
  representations in vector space,'' \emph{arXiv preprint arXiv:1301.3781},
  2013.

\bibitem{levy2014}
O.~Levy and Y.~Goldberg, ``Neural word embedding as implicit matrix
  factorization,'' in \emph{Advances in Neural Information Processing Systems},
  2014, pp. 2177--2185.

\bibitem{li2015}
Y.~Li, L.~Xu, F.~Tian, L.~Jiang, X.~Zhong, and E.~Chen, ``Word embedding
  revisited: A new representation learning and explicit matrix factorization
  perspective,'' 2015.

\bibitem{schmidhuber2015}
J.~Schmidhuber, ``Deep learning in neural networks: An overview,'' \emph{Neural
  Networks}, vol.~61, pp. 85--117, 2015.

\bibitem{zeiler2013}
M.~D. Zeiler, M.~Ranzato, R.~Monga, M.~Mao, K.~Yang, Q.~V. Le, P.~Nguyen,
  A.~Senior, V.~Vanhoucke, J.~Dean \emph{et~al.}, ``On rectified linear units
  for speech processing,'' in \emph{Acoustics, Speech and Signal Processing
  (ICASSP), 2013 IEEE International Conference on}.\hskip 1em plus 0.5em minus
  0.4em\relax IEEE, 2013, pp. 3517--3521.

\bibitem{werbos1990}
P.~J. Werbos, ``Backpropagation through time: what it does and how to do it,''
  \emph{Proceedings of the IEEE}, vol.~78, no.~10, pp. 1550--1560, 1990.

\bibitem{Bergstra_etal10}
J.~Bergstra, O.~Breuleux, F.~Bastien, P.~Lamblin, R.~Pascanu, G.~Desjardins,
  J.~Turian, D.~Warde-Farley, and Y.~Bengio, ``Theano: a {CPU} and {GPU} math
  expression compiler,'' in \emph{Proceedings of the Python for Scientific
  Computing Conference (SciPy)}, vol.~4, 2010.

\bibitem{cassidy2013}
A.~S. Cassidy, P.~Merolla, J.~V. Arthur, S.~K. Esser, B.~Jackson,
  R.~Alvarez-icaza, P.~Datta, J.~Sawada, T.~M. Wong, V.~Feldman, A.~Amir, D.~B.
  dayan Rubin, E.~Mcquinn, W.~P. Risk, and D.~S. Modha, ``Cognitive computing
  building block: A versatile and efficient digital neuron model for
  neurosynaptic cores,'' in \emph{in International Joint Conference on Neural
  Networks (IJCNN). IEEE}, 2013.

\bibitem{krishnan2005}
V.~Krishnan, S.~Das, and S.~Chakrabarti, ``Enhanced answer type inference from
  questions using sequential models,'' in \emph{Proceedings of the conference
  on Human Language Technology and Empirical Methods in Natural Language
  Processing}.\hskip 1em plus 0.5em minus 0.4em\relax Association for
  Computational Linguistics, 2005, pp. 315--322.

\bibitem{huang2008}
Z.~Huang, M.~Thint, and Z.~Qin, ``Question classification using head words and
  their hypernyms,'' in \emph{Proceedings of the Conference on Empirical
  Methods in Natural Language Processing}.\hskip 1em plus 0.5em minus
  0.4em\relax Association for Computational Linguistics, 2008, pp. 927--936.

\bibitem{Muller_Indiveri15}
L.~K. Muller and G.~Indiveri, ``Rounding methods for neural networks with low
  resolution synaptic weights,'' \emph{arXiv preprint arXiv:1504.05767}, 2015.

\bibitem{Wilson_Cowan72}
H.~Wilson and J.~Cowan, ``Excitatory and inhibitory interactions in localized
  populations of model neurons,'' \emph{Biophysical Journal}, vol.~12, pp.
  1--23, 1972.

\bibitem{Habenschuss_etal13}
S.~Habenschuss, Z.~Jonke, and W.~Maass, ``Stochastic computations in cortical
  microcircuit models,'' \emph{PLoS computational biology}, vol.~9, no.~11, p.
  e1003311, 2013.

\bibitem{Kappel_etal14}
D.~Kappel, B.~Nessler, and W.~Maass, ``Stdp installs in winner-take-all
  circuits an online approximation to hidden markov model learning,''
  \emph{PLoS computational biology}, vol.~10, no.~3, p. e1003511, 2014.

\bibitem{Neftci_etal13a}
E.~Neftci, S.~Das, B.~Pedroni, K.~Kreutz-Delgado, and G.~Cauwenberghs,
  ``Restricted boltzmann machines and continuous-time contrastive divergence in
  spiking neuromorphic systems,'' May 2013.

\bibitem{Neftci_etal14}
E.~Neftci, C.~Posch, and E.~Chicca, \emph{Neuromorphic Engineering}.\hskip 1em
  plus 0.5em minus 0.4em\relax UNESCO Encyclopedia of Life Support Systems,
  2014, ch.~23, (in press).

\bibitem{Diehl_Cook15}
P.~U. Diehl and M.~Cook, ``Unsupervised learning of digit recognition using
  spike-timing-dependent plasticity,'' \emph{Frontiers in Computational
  Neuroscience}, vol.~9, p.~99, 2015.

\bibitem{Abbott_Song99}
L.~Abbott and S.~Song, ``Asymmetric hebbian learning, spike timing and neural
  response variability,'' in \emph{Advances in Neural Information Processing
  Systems}, vol.~11, 1999, pp. 69--75.

\bibitem{Hussain_etal14}
S.~Hussain, A.~Basu, R.~M. Wang, and T.~J. Hamilton, ``Delay learning
  architectures for memory and classification,'' \emph{Neurocomputing}, vol.
  138, pp. 14--26, 2014.

\bibitem{Hochreiter_Schmidhuber97}
S.~Hochreiter and J.~Schmidhuber, ``Long short-term memory,'' \emph{Neural
  computation}, vol.~9, no.~8, pp. 1735--1780, 1997.

\bibitem{cho2014}
K.~Cho, B.~Van~Merri{\"e}nboer, C.~Gulcehre, D.~Bahdanau, F.~Bougares,
  H.~Schwenk, and Y.~Bengio, ``Learning phrase representations using rnn
  encoder-decoder for statistical machine translation,'' \emph{arXiv preprint
  arXiv:1406.1078}, 2014.

\bibitem{mikolov2014}
T.~Mikolov, A.~Joulin, S.~Chopra, M.~Mathieu, and M.~Ranzato, ``Learning longer
  memory in recurrent neural networks,'' \emph{arXiv preprint arXiv:1412.7753},
  2014.

\bibitem{le2015}
Q.~V. Le, N.~Jaitly, and G.~E. Hinton, ``A simple way to initialize recurrent
  networks of rectified linear units,'' \emph{arXiv preprint arXiv:1504.00941},
  2015.

\bibitem{mnih2014}
V.~Mnih, N.~Heess, A.~Graves \emph{et~al.}, ``Recurrent models of visual
  attention,'' in \emph{Advances in Neural Information Processing Systems},
  2014, pp. 2204--2212.

\bibitem{graves2013}
A.~Graves, A.-r. Mohamed, and G.~Hinton, ``Speech recognition with deep
  recurrent neural networks,'' in \emph{Acoustics, Speech and Signal Processing
  (ICASSP), 2013 IEEE International Conference on}.\hskip 1em plus 0.5em minus
  0.4em\relax IEEE, 2013, pp. 6645--6649.

\bibitem{boulanger2013}
N.~Boulanger-Lewandowski, Y.~Bengio, and P.~Vincent, ``Audio chord recognition
  with recurrent neural networks.'' in \emph{ISMIR}, 2013, pp. 335--340.

\bibitem{mikolov2012}
T.~Mikolov and G.~Zweig, ``Context dependent recurrent neural network language
  model.'' in \emph{SLT}, 2012, pp. 234--239.

\end{thebibliography}

\end{document}